\documentclass[11pt]{article}

\usepackage[preprint]{acl}

\usepackage{times}
\usepackage{latexsym}
\usepackage{amsmath}
\usepackage{amsfonts}
\usepackage[T1]{fontenc}
\usepackage[table,xcdraw]{xcolor}
\usepackage[utf8]{inputenc}

\usepackage{microtype}
\usepackage[skins,breakable]{tcolorbox}
\usepackage{listings}

\usepackage{graphicx}
\usepackage{algorithmic}
\usepackage{algorithm}
\usepackage{float}
\usepackage{subfig}
\usepackage{multirow}
\usepackage{booktabs}
\usepackage{graphicx}
\usepackage{enumitem}
\usepackage{colortbl,xcolor}

\newcolumntype{g}{>{\columncolor[HTML]{EAF0F6}}c}
\newcommand{\nece}[1]{#1}

\definecolor{latBlue}{HTML}{2B8CBE}   
\definecolor{latGreen}{HTML}{41AE76}  
\definecolor{latGray}{HTML}{B3B3B3}   

\title{GRACE: Reinforcement Learning for\\ \underline{G}rounded \underline{R}esponse and \underline{A}bstention under \underline{C}ontextual \underline{E}vidence}

\author{
  Yibo Zhao$^{1}$ \quad
  Jiapeng Zhu$^{1}$ \quad
  Zichen Ding$^{2}$ \quad
  Xiang Li$^{1}$\thanks{Corresponding Author: {xiangli@dase.ecnu.edu.cn}} \\
  $^1$School of Data Science and Engineering, East China Normal University \\
  $^2$Shanghai AI Laboratory \\
}

\newcommand{\ours}[0]{GRACE}

\begin{document}
\maketitle
\begin{abstract}
Retrieval-Augmented Generation (RAG) integrates external knowledge to enhance Large Language Models (LLMs), yet systems remain susceptible to two critical flaws: providing correct answers without explicit grounded evidence and producing fabricated responses when the retrieved context is insufficient.
While prior research has addressed these issues independently, a unified framework that integrates evidence-based grounding and reliable abstention is currently lacking. In this paper, we propose GRACE, a reinforcement-learning framework that simultaneously mitigates both types of flaws. 
GRACE employs a data construction method that utilizes heterogeneous retrievers to generate diverse training samples without manual annotation. A multi-stage gated reward function is then employed to train the model to assess evidence sufficiency, extract key supporting evidence, and provide answers or explicitly abstain. 
Experimental results on two benchmarks demonstrate that GRACE achieves state-of-the-art overall accuracy and strikes a favorable balance between accurate response and rejection, while requiring only 10\% of the annotation costs of prior methods. Our code is available at \url{https://github.com/YiboZhao624/Grace}.
\end{abstract}

\section{Introduction}


\begin{figure}[t]
    \centering
    \includegraphics[width=\linewidth]{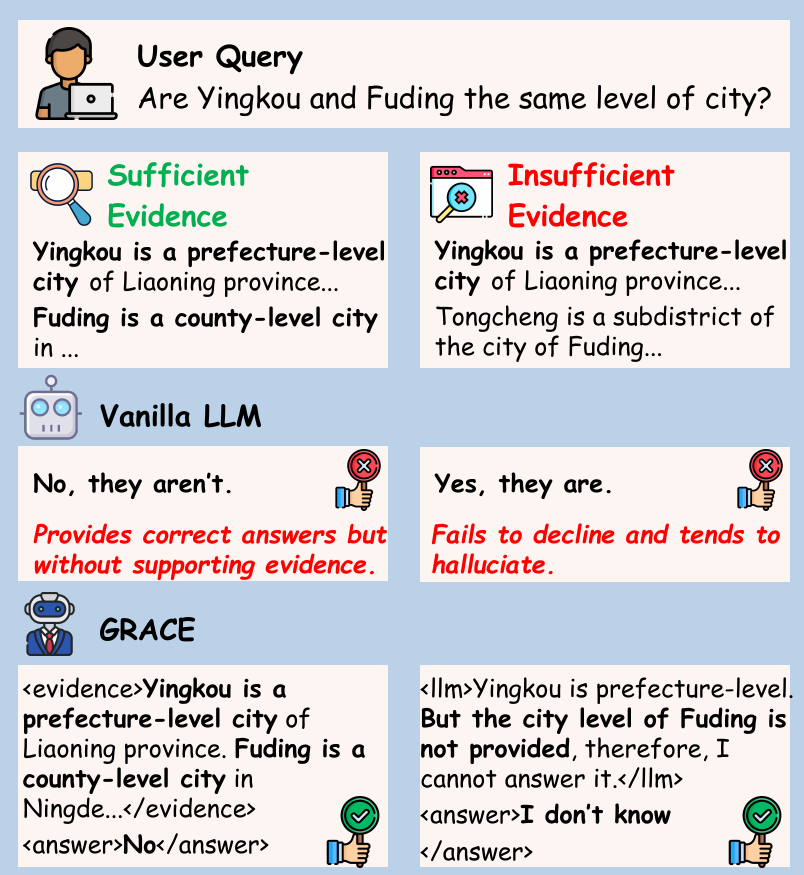}
    \caption{An example of responses generated by vanilla LLM and \ours\ under different retrieval results.}
    \label{fig:intro}
\end{figure}

With the rapid advancement, 
large language models (LLMs) have been widely applied in various fields, 
including machine translation~\cite{machine-translation-1, machine-translation-2}, knowledge-based question answering~\cite{knowledgeQA-1,knowledgeQA-2,knowledgeQA-3}, and mathematical reasoning~\cite{deepseek-math,deepseek-prover}. 
However, the knowledge contained in LLMs is limited, which is often insufficient to provide reliable responses for knowledge-intensive tasks~\cite{LLM-knowledge-boundary, LLM-knowledge-boundary-2}. 
To address this limitation, retrieval-augmented generation (RAG) integrates relevant context from external knowledge bases into the generative process of LLMs~\cite{RAG,RAG-2,self-rag}.

Nevertheless, hallucination remains a persistent issue in RAG systems~\cite{RAG-Halluciation-nocode, RAG-Halluciation-Agent}. As illustrated in Figure~\ref{fig:intro}, LLMs may either provide correct answers without grounding the key evidence (left), or produce fabricated responses despite lacking sufficient supporting evidence (right), 
thereby compromising the reliability of the generated output.~\cite{RAG-two-failures}. 
Therefore, building a trustworthy RAG system requires strengthening two key aspects: (1) improving evidential grounding, whereby the model provides correct answers while explicitly citing supporting evidence for transparency; and (2) fostering model honesty, enabling it to abstain from responding when the provided evidence is insufficient.

However, prior studies tend to strengthen these two aspects in isolation. One line of research leverages prompting~\cite{ground-prompt}, supervised fine-tuning (SFT)~\cite{ground-sft}, or reinforcement learning (RL)~\cite{groud-RL,ground-RL-2} to enhance grounding performance. 
Conversely, another branch of work focuses on calibrating refusal mechanisms through prompting~\cite{RAG-Abstentaion-prompt}, constraint decoding~\cite{RAG-Abstention}, or preference alignment~\cite{Trust-Align}.
The former often inadvertly encourges the model to speculate by prioritizing evidence localization over factual honesty, even in the absence of necessary information. 
In contrast, the latter tends to overlook the model’s proficiency in generating evidence-based answers. 
Consequently, these fragmented approaches fail to reconcile both capabilities simultaneously, compromising the overall trustworthiness of the RAG system. This naturally raises a research question: \emph{how can we simultaneously mitigate both types of hallucination while minimizing human annotation costs?} 

{To answer this question, we claim that an ideal RAG backbone should make the decision boundary explicit. 
The model should either produce an evidence-grounded response with explicit citations or, when the retrieved context is inadequate, provide a tentative answer accompanied by a clear disclaimer regarding the lack of supporting evidence.}
{Thus, 
we propose a sample-efficient framework \ours\ 
for training RAG backbones to reconcile evidence-integrated answering with reliable refusal.
Hence,
our framework comprises three key components: (1) a pipeline for constructing answerable and unanswerable training data from heterogeneous retrievers; (2) a multi-stage gated reward that provides multi-aspect outcome feedback; and (3) a modified Dynamic sAmpling Policy Optimization (DAPO) algorithm for effective fine-tuning.} 
To enhance generalization, we first curate a balanced training corpus leveraging heterogeneous retrievers to collect diverse candidate chunks. We then strategically partition queries into answerable and unanswerable categories by selectively withholding supporting evidence, thereby creating a benchmark for the model to learn evidence sufficiency.
During training, we utilize a multi-stage gated reward to jointly optimize the evidence grounding, answer accuracy, and abstention: a format reward ensures structural consistency, a path selection reward guides the model to judge evidence sufficiency (answering vs. abstaining), fostering model honesty, and a content accuracy reward encourages precise evidence referencing and correct final answers, improving evidential grounding.
Experimental results demonstrate that \ours~achieves state-of-the-art performance, outperforming significantly larger models and data-intensive baselines. Notably, with only 2,000 annotated samples, our 4B-parameter model not only surpasses existing benchmarks in overall accuracy but also achieves a superior synergy between answer correctness and refusal reliability, underscoring the exceptional sample efficiency of \ours.
Our contributions are threefold:

\begin{itemize}[leftmargin=*, itemsep=0pt]
\item \textbf{A retriever-based data construction pipeline} that curates diverse training samples, including both evidence-sufficient and evidence-insufficient scenarios. This enables the model to jointly optimize for accurate cited response generation and the robust capability to abstain.
\item \textbf{A joint optimization scheme} that simultaneously trains the model to assess retrieval sufficiency and to effectively utilize evidence, leading to state-of-the-art results on two datasets.
\item \textbf{New insights for training agentic RAG models}, revealing that sparse outcome-based rewards can lead to unintended behaviors in intermediate steps, especially in 
judging retrieval sufficiency. 
\end{itemize}

\section{Related Work}

Since the introduction of retrieval-augmented generation by ~\citep{RAG}, a number of subsequent studies have sought to improve RAG systems along various dimensions. In this work, we focus on three core aspects: answer accuracy, grounding fidelity, and abstention behavior.

To improve answer accuracy, a range of prompt-based techniques have been proposed~\cite{accuracy-prompt-1,accuracy-prompt-self-ask}, including workflows that employ summarization~\cite{accuracy-prompt-2-sure} and question decomposition~\cite{accuracy-prompt-decomposition}. While these methods are attractive due to their plug-and-play nature, they remain fundamentally constrained by the intrinsic capabilities of their backbone language models, which limit the degree of performance gain. 
Alternatives involving explicit training, such as supervised fine-tuning~\cite{accuracy-sft-RAFT} or RL~\cite{accuracy-RL-ARENA}, have also been explored. However, these training-based approaches typically overlook the issue of unanswerable or out-of-scope queries that frequently arise in real-world applications, reducing their practicality.
Efforts to strengthen grounding capability have given rise to frameworks like HalluGuard~\cite{evidence-grounding-rl-2}, which combines a large language model with a preference-optimized smaller model to verify retrieved evidence prior to answer generation. Similarly, TrustAlign~\cite{Trust-Align} constructs a large-scale preference dataset and applies Directedly Preference-Optimized training (DPO) in order to improve the alignment of outputs with trustworthy evidence. Despite the effectiveness, these methods often entail the generation or curation of vast amounts of training data, imposing a significant resource burden.
Regarding abstention, most existing approaches adopt prompt engineering~\cite{abstentation-prompt-1,abstention-prompt-2} or pipeline-based workflows~\cite{abstention-cooperate} to encourage the model to decline to answer when appropriate. Yet, these methods again depend heavily on the model’s inherent abstention capability.

Notably, existing methods often treat answer accuracy, evidence grounding, and abstention behavior as independent objectives, 
rendering the joint optimization of these dimensions within a unified framework an open challenge.



\section{Method}


\begin{figure*}[t]
    \centering
    \includegraphics[width=\textwidth]{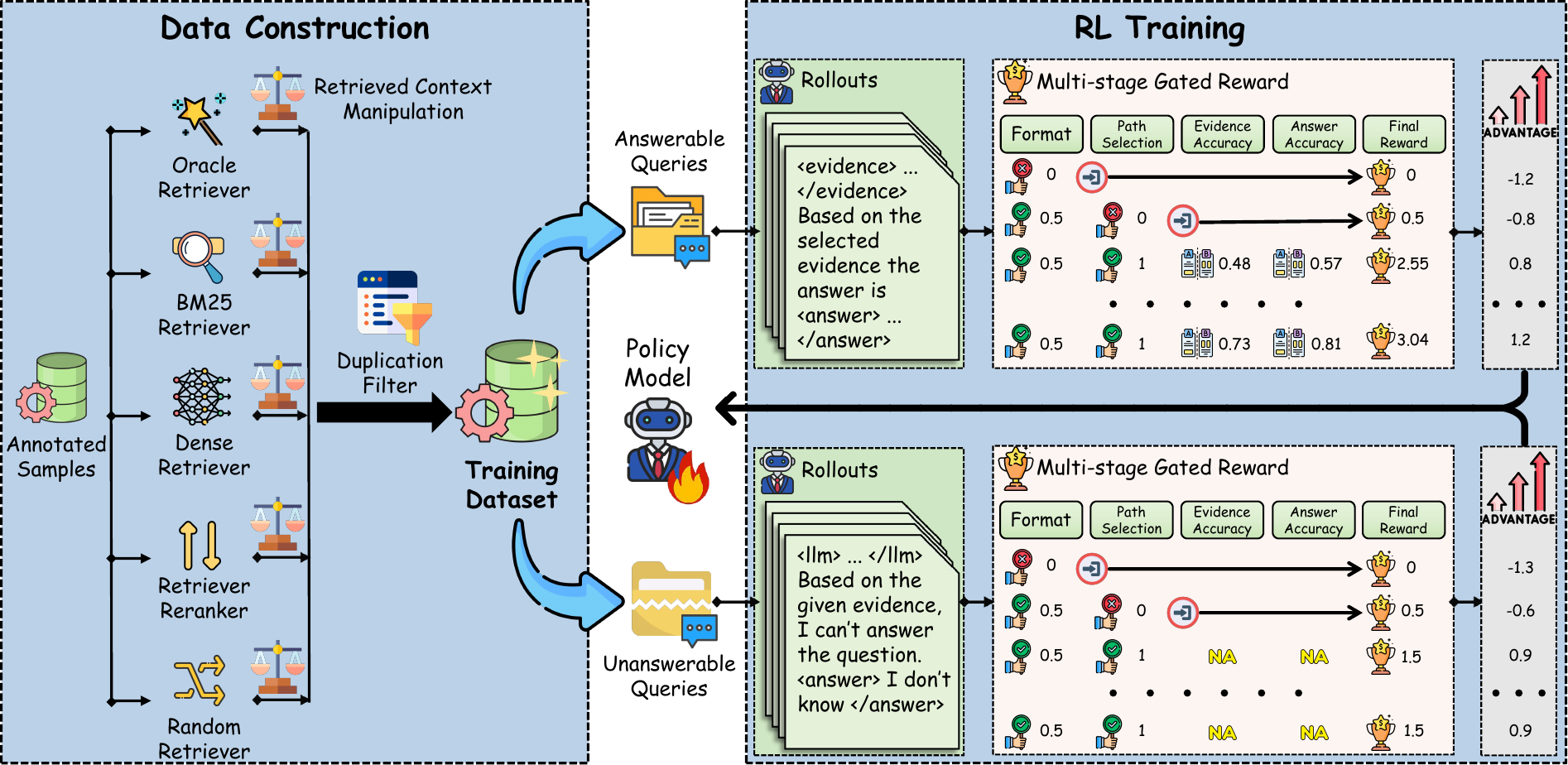}
    \caption{Overview of the full pipeline of our proposed method.}
    \label{fig:main}
\end{figure*}

Our framework comprises three integral components: {data construction}, {reward function design}, and {training algorithm}. 
To facilitate the subsequent discussion, we first formalize our problem setting and notations.
With the annotated dataset $\mathcal D$, given the $i$-th query $q^i$, the knowledge base contains a set of ground-truth supporting passages $\mathcal C_\text{gt}^i$, 
and a set of distracting passages $\mathcal C_\text{dis}^i$. Crucially, each ground-truth passage contains specific key evidence sentences or phrases, denoted as $e_\text{gt}^i$, which are essential for answering the query and serve as the reference for fine-grained verification. 
Additionally, each query is associated with a ground-truth answer $a^i_\text{gt}$. In conclusion, a dataset sample is represented as $s^i = \{q^i, \mathcal C_\text{gt}^i, \mathcal C_\text{dis}^i, a^i_\text{gt}, e^i_\text{gt}\} \in \mathcal D$. 
In the RAG framework, a retriever fetches the top-$k$ candidate passages $\mathcal C_\text{ret}^i = \{c_1^i,\cdots,c_k^i\}$ for each query $q^i$ from the knowledge base. We then format the query and retrieved passages with a prompt template $\mathcal P(\cdot, \cdot)$ to construct the prompt $p^i = \mathcal P(q^i,\mathcal C_\text{ret}^i)$. Finally, the LLM $\pi_\theta$ generates an output $o^i$ conditioned on the prompt, i.e., $o^{i} = \pi_\theta(p^i)$.


\subsection{Data Construction}\label{method:data}


Existing RAG datasets~\cite{hotpotqa, QASPER} predominantly focus on scenarios where provided evidence is sufficient to answer a query, i.e., $\mathcal C_\text{gt}^i \subseteq \mathcal C_\text{ret}^i$. This positive-only bias overlooks the critical need for models to evaluate evidence sufficiency, hindering their ability to develop a reliable refusal mechanism when the retrieved context is irrelevant or incomplete.

A straightforward approach to constructing a dataset containing both evidence-sufficient and evidence-insufficient samples is to randomly split $\mathcal D$ into two subsets. In the first subset, $q^i$ is paired with $\mathcal C^i_\text{gt}$ to simulate sufficient evidence scenarios, while in the second subset, $q^i$ is paired with $\mathcal C_\text{dis}^i$ to create insufficient evidence scenarios.
However, this naive construction has notable drawbacks. The evidence-sufficient samples contain only ground-truth evidence without any distractors, significantly reducing the difficulty of the task. Conversely, because the insufficient samples are randomly selected from $\mathcal C_\text{dis}^i$, they may lack relevance to the query. This makes it trivial for the model to distinguish between sufficient and insufficient evidence, \nece{encouraging the learning of superficial patterns rather than truly enhancing the model's ability to discern evidence sufficiency}.
To address this, we propose a data construction framework that leverages heterogeneous retrievers to systematically synthesize a balanced training set that forces the model to distinguish between answering and abstention.

Let $\mathcal R$ denote a set of retrievers. 
For each query $q^i$ and each retriever $r \in \mathcal R$, we retrieve a top-$k$ context set $\mathcal C_\text{ret}^{i, (r)}$, yielding $| \mathcal R |$ context variants per annotated sample.
For each retriever $r$, we obtain a retriever-specific training set $\mathcal D^{(r)}$. \nece{Notably, the distribution of retrieved samples is often skewed, as the retrieved set is typically dominated by candidates that contain key evidence. This class imbalance poses a substantial challenge for training models to assess evidence sufficiency accurately.} 
To mitigate this, we randomly split $\mathcal D^{(r)}$ into two equally-sized subsets: evidence sufficient subset $\mathcal D^{(r)}_\text{suff}$ and evidence insufficient subset $\mathcal D^{(r)}_\text{insuff}$. To promote diversity in the training data, we employ retriever-specific random seeding.
Building on the split, we construct the final context for each annotated sample by manipulating the retrieved passages while keeping the context length fixed at $k$ and preserving the retrieval order of the remaining passages.
Specifically, for $s^i \in \mathcal D^{(r)}_\text{suff}$, we enforce the presence of supporting evidence: if $\mathcal C^{i,(r)}_\text{ret}$ does not contain the ground-truth supporting passage(s) $\mathcal C_\text{gt}^{i}$, we replace the lowest-ranked passage(s) with $\mathcal C_\text{gt}^{i}$, leaving the relative order of the other passages unchanged. For $s^i \in \mathcal D^{(r)}_\text{insuff}$, we simulate insufficient-evidence scenarios by strictly excluding ground-truth support from the context. We remove any retrieved passage that matches $\mathcal C_\text{gt}^i$ and replace it with distracting passages from $\mathcal C_\text{dis}^{i}$, again preserving the set size.
Finally, we aggregate all the retriever-specific datasets $\{\mathcal D^{(r)}\}_{r\in\mathcal R}$ into a single training corpus with a duplication filter. 
We deduplicate the dataset by retaining only unique pairs of queries $q^i$ and retrieved contexts $\mathcal C_\text{ret}^{i,(r)}$. 
The merged dataset is used as our final training set $\mathcal D_\text{train}$ and $\mathcal D_\text{train} = \mathcal D_\text{suff} \cup \mathcal D_\text{insuff}$.

{Through the deployment of heterogeneous retrievers, we ensure that the retrieved distractor chunks exhibit varying degrees of semantic relevance to the query. This diverse context forces the model to discern evidence sufficiency across multiple levels of difficulty. 
Furthermore, by manipulating the retrieved context, we maintain an equitable distribution between evidence-sufficient and evidence-insufficient instances. 
This strategic balancing prevents the model from falling into reward hacking, where it might otherwise over-fit to a single decision path to maximize cumulative rewards. 
Finally, we apply a deduplication filter to refine the dataset, guarding against redundant samples that could skew the model's learning process.}

\subsection{Multi-stage Gated Reward}

{As stated in the introduction, an ideal RAG backbone model should have the following properties. 
First, it must explicitly indicate whether the retrieved evidence is sufficient, enabling users to judge whether the answer should be trusted. 
Second, when the model deems the evidence sufficient, it should identify the specific sentence(s) on which the answer is grounded. 
Finally, when the model considers the evidence insufficient, it should still provide a best-effort answer based on its parametric knowledge, while issuing a clear disclaimer that the response may be susceptible to errors.
To this end, we design an XML-style response template and hope our trained model can internalize the following behaviors.
If the model outputs an <evidence> tag, it indicates that the model judges the evidence to be sufficient and includes the key supporting sentence(s) within the tag. 
If the model outputs an <llm> tag, it indicates that the model judges the evidence to be insufficient and therefore relies on its internal knowledge to infer an answer, accompanied by a reliability disclaimer. 
The model always outputs an <answer> tag to highlight the final prediction.}

Our reward function is explicitly designed to satisfy three essential criteria: (1) assessing feasibility of answering, i.e., whether the retrieved context contains sufficient evidence to answer the query, (2) identifying the specific supporting evidence when available, and (3) adaptively producing an evidence-grounded answer, or providing a best-effort answer while explicitly warning that the retrieved evidence is insufficient. 
Guided by these criteria, we adopt a multi-stage gated reward structure comprising format, path, and content components. 
The total reward $R(o)$ is the cumulative sum of these components, calculated sequentially. 

\textbf{Format Reward} ($R_\text{f}$): To facilitate structured parsing for rule-based reward calculation, we introduce XML-like special tokens. The output must strictly adhere to one of two structures: either an \texttt{<evidence>}...\texttt{</evidence>} block (indicating sufficient evidence) or an \texttt{<llm>}...\texttt{</llm>} block (indicating insufficient evidence and leveraging internal parametric knowledge), followed by an independent \texttt{<answer>}...\texttt{</answer>} block. 
We define a binary format reward $R_\text{f}$: the model receives a fixed reward of $0.5$ if the output strictly complies with these constraints, and $0$ otherwise. If $R_\text{f}=0$, the evaluation terminates immediately, and the total reward is set to $0$. 
{By early-stopping the reward accumulation upon format failure, we prevent the model from spurious content rewards through structurally non-compliant outputs, and enforce strict adherence to the predefined schema.}

\textbf{Path Selection Reward} ($R_\text{p}$): For outputs adhering to the valid format, we evaluate the model's decision path. For {answerable, evidence sufficient} samples $s^i \in \mathcal D_\text{suff}$, the model is expected to invoke the \texttt{<evidence>} tag to utilize retrieved contexts; for {unanswerable, evidence insufficient} samples $s^i \in \mathcal D_\text{insuff}$, it should generate the \texttt{<llm>} tag, indicating the insufficiency of retrieved evidence and reliance on internal parametric knowledge. 
A correct path selection yields $R_p = 1$. Conversely, an incorrect path results in $R_p = 0$, at which point the episode terminates with a cumulative reward $R(o) = R_p + R_f = 0.5$. 

\textbf{Content Accuracy Reward} ($R_\text{c}$): Upon a correct selection of the \texttt{<evidence>} path, we evaluate the fidelity of the extracted evidence $o_\text{evid}$ within \texttt{<evidence>} block, and the precision of the final answer $o_\text{ans}$ within \texttt{<answer>} block. We compute the Rouge-L F1 score for both components against their respective ground truths. The content accuracy reward is defined as a weighted sum: $R_\text{c} = \alpha\cdot\text{Rouge-L}_\text{F1}(o_\text{evid},e_\text{gt}) +\beta\cdot \text{Rouge-L}_\text{F1}(o_\text{ans},a_\text{gt})$, where $\alpha$ and $\beta$ adjust the relative importance of evidence extraction and answer accuracy. In cases with multiple references, we utilize the maximum score across all candidates. If the model correctly selects the \texttt{<llm>} path, $R_c$ is not computed. While the model may optionally provide an answer via parametric knowledge, it remains unscored to disincentivize reliance on memorization and prioritize evidence-grounded reasoning. 

We formulate the total reward $R(o)$ as a step function to enforce prerequisite constraints:
\begin{equation}
\small 
R(o) = 
\begin{cases} 
0 & \text{invalid format}\\
R_f & \text{valid format, wrong path} \\
R_f + R_p & \text{correct \texttt{<llm>} path} \\
R_f + R_p + R_c & \text{correct \texttt{<evidence>} path}
\end{cases}
\end{equation}

{This hierarchical design trains the model at two levels: (i) Decision Calibration, which mandates the model to assess evidence sufficiency and abstain when necessary; and (ii) Execution Accuracy, which ensures that for answerable queries, the model generates precise responses grounded in the provided context.}
To operationalize this multi-stage gated reward, we provide the reward computation in pseudocode in Algorithm~\ref{alg:reward}.

\subsection{RL Training Stage}

Our approach builds upon DAPO~\cite{dapo}, an existing on-policy reinforcement learning algorithm. Given an input prompt $p^i$, the LLM policy $\pi_\theta$ samples a group of $G$ responses $\{o_j^i\}_{j=1}^G$. Each response is evaluated by the reward function $R(\cdot)$ defined previously, yielding a scalar reward $r_j^i = R(o_j^i)$. Subsequently, we calculate the group-wise advantage as:
\begin{equation}
    \hat A_j^i = \frac{r_j^i - \text{mean}\left(\left\{r_l^i\right\}_{l=1}^G\right)}{\text{std}\left(\left\{r_l^i\right\}_{l=1}^G\right)}
\end{equation}
Consistent with DAPO, this advantage score is assigned to every token within the sequence, i.e., $A^i_{j,t} = A(o^i_j), \forall t \in [1,2,\cdots, \text{len}(o_j^i)]$.

For each token, we define the policy ratio $\rho^i_{j,t}$ as:
\begin{equation}
    \rho^i_{j,t} = \frac{\pi_\theta\left(o^i_{j,t} \mid p^i, o^i_{j,<t}\right)}{\pi_\text{ref}\left(o^i_{j,t} \mid p^i, o^i_{j,<t}\right)},
\end{equation}
which quantifies the deviation of the current policy from the reference model. 
Following DAPO, we adopt the {clipped surrogate objective with the clip-higher strategy} to ensure update stability while encouraging exploration. The token-level objective is formally defined as:
\begin{equation}
    l^i_{j,t} = \min \left(  \rho^j_{j,t}A^i_{j,t},
    \text{clip} \left ( \rho^i_{j,t} , 1 - \epsilon_\text{l}, 1+\epsilon_\text{h} \right )A^i_{j,t}\right)
\end{equation}\label{eq:l}
The asymmetric clipping function bounds the policy ratio, enforcing conservative updates while permitting larger probability increases for initially low-probability tokens. 
Further, we omit the KL penalty term and utilize a token-level policy gradient loss to normalize against varying response lengths within a group. The total loss $\mathcal L$ is formulated as:
\begin{equation}
\begin{aligned}
\mathcal L = \mathbb E_{e^i\in \mathcal D_\text{train}, \{o^i_j\}_{j=1}^G \sim \pi_\text{ref}(\cdot \mid p^i)} \\
\left[\frac{1}{\sum_{j=1}^G |o^i_j|} \sum_{j=1}^G \sum_{t=1}^{|o_j^i|} l_{j,t}^i\right]
\end{aligned}
\end{equation}

Notably, our implementation diverges from the original DAPO by excluding dynamic sampling. In the original DAPO, dynamic sampling is essential to handle binary feedback, as groups with identical rewards yield zero advantage and thus contribute no learning signal. In contrast, our reward function returns continuous floating-point values. This continuous nature ensures reward diversity within each group, guaranteeing valid advantage estimation and effective gradient updates without sampling.

\section{Experiment}

\begin{table*}[ht]
\centering
\caption{Main Results on QASPER and HotpotQA, where the evidences are retrieved by Qwen3-Embedding-0.6B with $k=3$. The \textbf{best results} are bold, and the \underline{runner-up results} are underlined. The number of answerable and unanswerable questions in each dataset is denoted as {({answerable}: {unanswerable})}. Methods with $\dagger$ sign come from the released checkpoints by the corresponding authors.}
\label{tab:main_results_tab}
\resizebox{\textwidth}{!}{\begin{tabular}{lcccccggcccccgg}
\toprule
\multirow{3}{*}{} & \multicolumn{7}{c}{\textbf{QASPER} \small{(1200 : 251)}} & \multicolumn{7}{c}{\textbf{HotpotQA}\small (296:204)} \\
\cmidrule(lr){2-8} \cmidrule(lr){9-15} 
& \multicolumn{3}{c}{Answerable} & \multicolumn{2}{c}{Unanswerable} & \multicolumn{2}{c}{\textbf{Overall}} & \multicolumn{3}{c}{Answerable} & \multicolumn{2}{c}{Unanswerable} & \multicolumn{2}{c}{\textbf{Overall}} \\
\cmidrule(lr){2-4} \cmidrule(lr){5-6} \cmidrule(lr){7-8} \cmidrule(lr){9-11} \cmidrule(lr){12-13} \cmidrule(lr){14-15}
& EM & F1 & LJ & Acc. & LJ & \textbf{Acc.} & \textbf{B. Acc.} & EM & F1 & LJ & Acc. & LJ & \textbf{Acc.} & \textbf{B. Acc.}\\
\midrule
\rowcolor[HTML]{F2F2F2}
\multicolumn{15}{l}{\textbf{Prompt based}} \\
Qwen3-4B          
& 0.0  & 50.23 & \textbf{78.83} & 5.58 & 42.23 & \underline{72.50} & 60.53
& 17.57 & 54.31 & 89.53 & 16.18 & 60.78 & \underline{77.80} & \underline{75.16}\\
Llama3-8B-Instruct 
& 1.33 & 54.60 & 70.58 & 12.75 & 30.68 & 63.68 & 50.63
& 8.78 & 39.15 & 27.36 & 48.53 & \underline{89.22} & 52.60 & 58.29\\
DeepSeek-chat     
& 0.0  & 51.88 & 69.33 & 50.60 & 72.11 & 69.81 & 69.57
& 33.45 & 60.46 & 68.92 & \underline{75.00} & 75.98 & 71.80 & 72.45\\
SuRe on Qwen3-4B  
& 5.92  & 52.43 & 46.17 & 0.0 & 0.40 & 38.25 & 23.28
& 26.35 & 63.62 & 65.54 & 0.0 & 1.47 & 39.40 & 33.51\\
SuRe on Llama3-8B-Instruct 
& 5.25  & 55.01 & 54.83 & 0.0 & 1.59 & 45.62 & 28.21
& 35.14 & 69.18 & 83.11 & 0.0 & 0.0 & 49.20 & 41.55 \\
SuRe on DeepSeek-chat 
& 13.08 & 58.68 & 57.92 & 0.0 & 7.17 & 49.14 & 32.54
& 51.69 & 78.49 & 84.12 & 0.0 & 0.98 & 50.20 & 42.55\\
\midrule
\rowcolor[HTML]{F2F2F2}
\multicolumn{15}{l}{\textbf{SFT based}} \\
Qwen3-4B
& \textbf{28.92} & 69.53 & 62.25 & 67.33 & 67.33 & 63.13 & 64.79
& 64.53 & 84.77 & 84.12 & 44.12 & 44.12 & 67.80 & 64.12\\
Llama3-8B-Instruct
& \underline{27.92} & 70.21 & 62.75 & 62.95 & 62.95 & 62.78 & 62.85
& \textbf{67.23} & \underline{86.38} & 87.50 & 49.51 & 50.49 & 72.40 & 69.00\\

\midrule
\rowcolor[HTML]{F2F2F2}
\multicolumn{15}{l}{\textbf{RL based}} \\
TrustAlign-3B$^\dagger$ 
              & 0.0   & 41.20 & 41.17 & \underline{78.09} & \underline{80.88} & 48.04 & 61.03
              & 0.0   & 38.66 & 44.59 & 55.88 & 67.16 & 53.80 & 55.88\\
TrustAlign on Qwen3-4B
              & 0.0   & 44.50 & 36.83 & 0.0   & 60.96 & 41.01 & 48.90
              & 0.0   & 32.27 & 11.49 & 0.0   & \textbf{96.57} & 46.20 & 54.03\\
TrustAlign-7B$^\dagger$ 
              & 0.0   & 43.63 & 33.67 & 0.0   & 88.45 & 43.14 & 38.41
              & 0.0   & 34.02 & 24.32 & 0.0   & 86.76 & 49.80 & 55.54\\
TrustAlign-8B$^\dagger$ 
              & 0.08  & 34.72 & 9.67  & \textbf{97.61} & \textbf{97.61} & 24.88 & 53.64
              & 0.0   & 30.57 & 22.64 & \textbf{88.73} & 88.73 & 49.60 & 55.69 \\
ARENA on Qwen3-4B 
              & 23.25 & 68.08 & 73.50 & 19.92  & 20.72 & 64.37 & 47.11
              & 64.86 & 86.05 & \underline{90.54} & 38.73  & 38.73  & 69.40 & 64.64\\
ARENA-7B$^\dagger$ 
              & 22.50 & 69.73 & 69.67 & 4.78  & 7.97  & 58.99 & 38.82
              & \underline{65.54} & \underline{86.27} & 88.18 & 5.88  & 9.31  & 55.60 & 48.75\\
ARENA-8B$^\dagger$ 
              & 23.58 & 68.66 & 69.17 & 0.0   & 0.40  & 57.27 & 34.79
              & 64.53 & 85.43 & 87.84 & 0.0   & 0.0   & 52.00 & 43.92\\
\midrule
\rowcolor[HTML]{F2F2F2}
\multicolumn{15}{l}{\textbf{Agentic Models}} \\
R1-Searcher-7B$^\dagger$
              & 15.42 & 59.80 & 48.83 & 17.13 & 19.92 & 43.83 & 34.38
              & 51.35 & 74.59 & 69.59 & 65.20 & 65.20 & 67.80 & 67.40 \\
Search-R1-7B$^\dagger$
              & 18.50 & 64.76 & 61.00 & 30.68 & 33.07 & 56.17 & 47.03
              & 56.08 & 79.49 & 78.72 & 47.55 & 47.55 & 66.00 & 63.13 \\
SimpleDeepSearcher-7B$^\dagger$
            & 22.92 & 67.13 & 67.17 & 16.73 & 17.93 & 58.65 & 42.55
            & 45.61 & 68.47 & 63.18 & 64.71 & 64.71 & 63.80 & 63.94\\
\midrule
\rowcolor[HTML]{F2F2F2}
\multicolumn{15}{l}{\textbf{Our methods}}\\
GRACE-Qwen3-4B  
& 27.08 & \textbf{73.09} & \underline{74.25} & 74.90 & 74.90 & \textbf{74.36} & \textbf{74.58}
& 65.20 & 84.95 & \textbf{91.55} & 66.18 & 66.18 & \textbf{81.20} & \textbf{78.87}\\
GRACE-Llama3.1-8B 
& 27.08 & \underline{72.02} & 70.08 & 74.50 & 74.50 & 70.85 & \underline{72.29}
& 65.20 & \textbf{86.40} & 88.85 & 42.16 & 44.61 & 70.80 & 66.73\\
\bottomrule
\end{tabular}}
\end{table*}

This section provides the main experimental results. Due to space limitations, we defer additional results to the Appendix, including: ablation study (Appendix~\ref{app:ablation-study}), performance on different retrievers (Appendix~\ref {app:retriever}), comparison of top-$k$ variants (Appendixheizen~\ref{app:top-k}),  performance on out-of-distribution data (Appendix~\ref{app:ood}), impact of training on general capabilities (Appendix~\ref{app:general}), and case study (Appendix~\ref{app:case-study}). 

\subsection{Experiment Settings}

Our experiments are conducted on two knowledge-intensive QA datasets: QASPER~\cite{QASPER} and HotpotQA~\cite{hotpotqa}.
We use two backbone models, Qwen3-4B~\cite{qwen3} and Llama3.1-8B-Instruct~\cite{llama3.1}. For both datasets, we follow the original train-test splits to ensure reproducibility. For HotpotQA, we sample 2,000 training instances and 500 test instances. For QASPER, we use the full dataset, consisting of 2,593 training entries and 1,451 test entries. 
Additional implementation details, including baseline configurations, prompt templates, and hyperparameter setup, are provided in Appendix~\ref{app:settings}.


\subsection{Baselines}

We consider four types of baselines.
\textbf{(1) Prompt-based methods.} This group includes direct prompting and SuRe~\cite{accuracy-prompt-2-sure}, applied to Qwen3-4B, Llama3.1-8B-Instruct and DeepSeek-Chat.
\textbf{(2) SFT-based methods.} We perform sft on Qwen3-4B and Llama3.1-8B-Instruct with our constructed training data ({details are given in Appendix~\ref{app:sft-data}}). 
\textbf{(3) RL-based methods} comprise ARENA~\cite{accuracy-RL-ARENA} and TrustAlign~\cite{Trust-Align}. 
\textbf{(4) Agentic methods} include SimpleDeepSearcher~\cite{simpledeepsearcher}, R1 Searcher~\cite{r1-searcher}, and Search-R1~\cite{searchr1}.

To ensure a fair and faithful comparison, we evaluate all baselines with their original inference templates from their respective papers. On top of these baseline-specific templates, we prepend a {\emph{unified three-shot demonstration}} containing two answerable examples with correct answers and one unanswerable example with a refusal-style response.
For agentic RAG baselines, we directly feed our retrieved chunks to the model. If the model executes an \texttt{<answer>} action, we treat it as indicating that the evidence is sufficient; if it instead executes a \texttt{<search>} action, we treat it as abstaining. To maintain consistency with our experimental setup, all methods are limited to a single interaction turn.

\subsection{Metrics}

To evaluate the model performance comprehensively, we divide the test set into two subsets: \textit{answerable} and \textit{unanswerable}, depending on whether the ground-truth evidence chunks are successfully retrieved. 
For answerable queries, we assess answer quality using Exact Match (EM), BERT F1 Score (F1) by \texttt{bert-baes-uncased}, and accuracy of LLM-as-a-judge (LJ) by \texttt{DeepSeek-V3.2-chat}. 
For unanswerable queries, we examine the model’s ability to follow instructions and refuse responses, evaluated by EM and LJ, respectively. Notably, we prioritize the presence of a reliability disclaimer over the factual correctness of the answer; a response is deemed incorrect if the model fails to signal evidence insufficiency, even if it happens to guess the correct answer using its internal knowledge.
To quantify performance, we report Accuracy (Acc.) to reflect the model's effectiveness under a real-world distribution, and Balanced Accuracy (B. Acc.) to evaluate its ability to answer and abstain with equal importance. 

\subsection{Main Results}

\begin{figure*}[t]
    \centering
    \includegraphics[width=\textwidth]{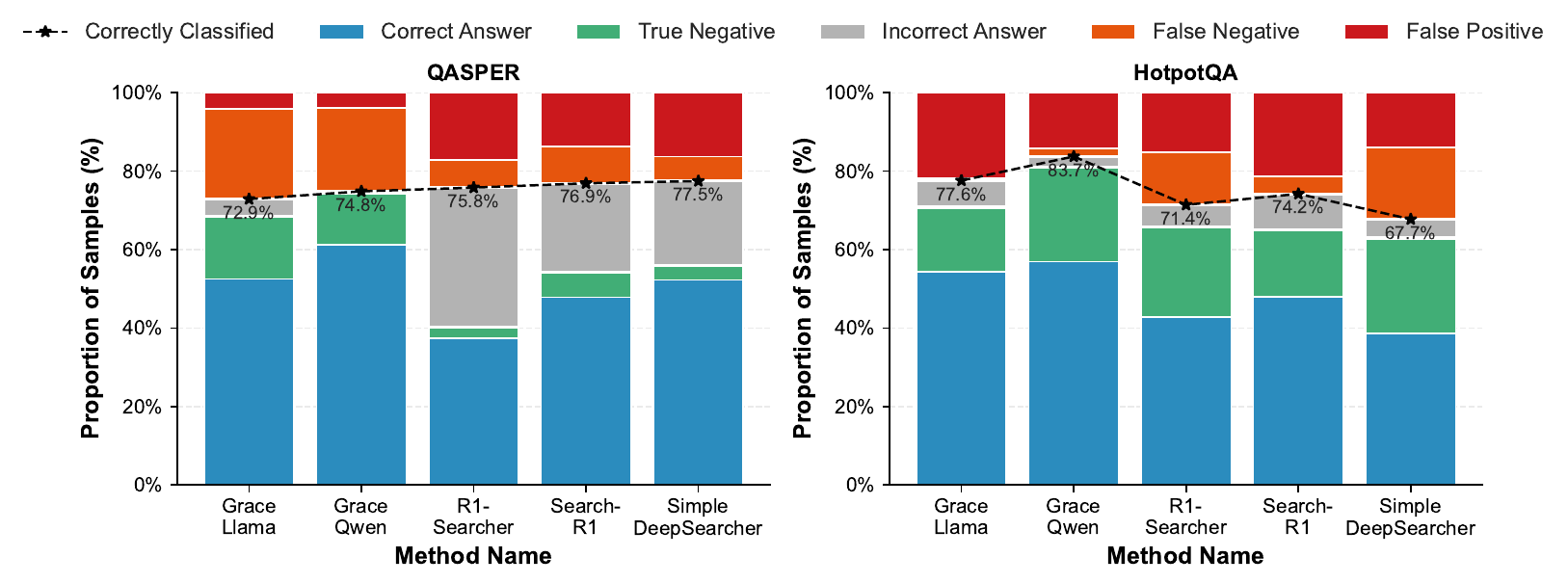}
    \caption{Error analysis of the proposed method versus agentic RAG baselines. 
    Classification accuracy is indicated by the ``Correctly Classified'' line.
    Proportions are calculated by the average results among different retrievers.}
    \label{fig:error}
\end{figure*}

As shown in Table~\ref{tab:main_results_tab}, our method \ours-Qwen3-4B model achieves the best overall performance in both accuracy and balanced accuracy across two datasets, 
which can 
effectively balance precise answering and appropriate refusal.
Specifically, compared with the untrained base model, GRACE improves the unanswerable LJ score from 42.23\% to 74.90\% on QASPER and
the answerable EM score from 17.57\% to 65.20\% on HotpotQA.
Similar trends are also observed in 
GRACE-Llama3.1-8B model.
All these results 
validate the generalization and effectiveness of our approach across different backbone architectures.



Regarding prompt-based methods, it is noteworthy that they can achieve surprisingly competitive performance when provided with appropriate prompts. For example, Qwen3-4B achieves runner-up overall accuracy across both benchmarks;
Llama3-8B-Instruct's unanswerable LJ score is 89.22\% on HotpotQA, which is the runner-up;
DeepSeek-chat ranks second in the unanswerable accuracy on HotpotQA. However, when injected with more sophisticated prompting strategies such as SuRe, we observe a drastic collapse in refusal capabilities for unanswerable queries. For instance, DeepSeek-chat's refusal accuracy on HotpotQA plummets from 75.98\% to a mere 0.98\%. We attribute this failure to SuRe's inherent assumption that the retrieved evidence is necessarily sufficient to support an answer, leaving little room for the model to acknowledge information missing.

For SFT-based approaches, we observe substantial improvements in answerable EM, F1 scores, and unanswerable accuracy over base models after training. For instance, the SFT-trained Llama3.1-8B-Instruct achieves the highest EM score and the runner-up F1 score on HotpotQA, indicating that our constructed data effectively guides the model to adhere to the required output formats. 
Conversely, their performance on the LJ metric exhibits instability. 
We attribute this to the absence of Chain-of-Thought (CoT) reasoning in our training data, which likely leads the model to superficially memorize answer templates rather than internalize the underlying reasoning process needed to solve the problem~\cite{cot-sft, cot-sft-2}.
When examining RL-based approaches, we find 
TrustAlign and ARENA exhibit two diverging behaviors. 
While TrustAlign achieves a near-perfect refusal accuracy (96.57\%), it struggles to correctly address answerable queries. 
This phenomenon suggests a case of over-alignment, where the model becomes overly conservative and reluctant to respond. 
In contrast, ARENA performs well on answerable queries but fails on unanswerable ones. This is primarily because its reward mechanism fails to account for scenarios where evidence is missing; consequently, the model resorts to reward hacking, attempting to fabricate an answer even when a refusal is warranted, solely to maximize its reward signal.
These observations suggest that both models are specialists rather than generalists.


\subsection{Comparison with Agentic Models}
We also incorporate recently prominent Agentic RAG models SimpleDeepSearcher-7B, R1-Searcher-7B and Search-R1-7B, treating their evaluation as a verification of single-step performance within the RAG context. 
Despite their sophisticated reasoning and tool-use capabilities, these models exhibit worse performance in our single-turn scenarios than \ours.
Notably, their overall performance significantly lags behind our method, even the strongest agentic baseline trails our GRACE-Qwen3-4B by more than 15\% in overall accuracy across both datasets (see Table~\ref{tab:main_results_tab}).

To investigate this discrepancy, we formulate evidence sufficiency identification as a binary classification task, where Positive (P) denotes sufficient evidence and Negative (N) denotes insufficient evidence. We categorize the results into True Positive (TP), True Negative (TN), False Positive (FP), and False Negative (FN). Within the TP group, we further distinguish between Correct (C) and Incorrect (I) answers, such that $TP = C + I$.
To quantify this, we define classification accuracy as the success rate in identifying evidence sufficiency, $\frac{TP+TN}{\text{Total}}$, and answer precision $\frac{C}{C+I}$.

As illustrated in Figure~\ref{fig:error}, on QASPER, while baseline agentic models achieve classification accuracy comparable to \ours\ (see black dashed curve), their answer precision is significantly lower. We attribute this to a domain mismatch: these models are typically trained on general open-domain data, whereas QASPER focuses on specialized academic QA, where identifying evidence sufficiency is relatively straightforward, but generating technically accurate answers is more challenging. Moreover, these baselines exhibit a high False Positive Rate (FPR), indicating a tendency toward overconfidence.
Conversely, on HotpotQA, the multi-hop, open-domain benchmark, these models exhibit a notably weaker capacity for evidence assessment compared to \ours, where agentic models score approximately 70\%, whereas our GRACE-Qwen3-4B achieves 83.7\%. 
We attribute this deficiency to the limitations of sparse outcome rewards used in GRPO-based training. Without dense, step-wise supervision, a model may receive positive reinforcement for a correct final answer even if its intermediate reasoning step is flawed. 
This supports the view that stronger performance stems from dense and reliable verification signals, rather than sparse outcome-based rewards, echoing the shift from DeepSeek-Math~\citep{deepseek-math} to DeepSeek-Math-V2~\cite{deepseek-math-v2}.


\section{Conclusion}

In this paper, we presented \ours, an RL framework dedicated to simultaneously developing reliable and transparent RAG backbone models. 
We proposed a pipeline to construct answerable and unanswerable training samples from heterogeneous retrievers.
Then, we designed a multi-stage gated reward mechanism that provides multi-aspect feedback, empowering models to discern evidence sufficiency and perform explicit grounding or informed refusal.
Finally, we proposed a modified DAPO algorithm to ensure training stability. Empirical results demonstrate that \ours\ outperforms all the baselines and achieve a superior synergy between answer correctness and refusal reliability.

\section*{Limitations}

Despite the promising results, our work has two main limitations.
First, constrained by computational resources, we could not extend our evaluation to larger models or conduct multiple runs to mitigate the impact of experimental randomness. 
While our method proves effective on smaller scales, verifying its scalability and performance on models with significantly larger parameters remains an important direction for future investigation.
Second, our approach currently relies on the availability of ground-truth key evidence for each query during training. This dependency on fine-grained annotations limits our ability to test on a broader range of datasets where such detailed labels are absent. Exploring methods to reduce this reliance, for instance, through weak supervision or automated evidence extraction, will be a primary focus of our future work.

\section*{Ethical Consideration}

\paragraph{Potential Risks.} Our model is designed for academic research in information retrieval and reasoning. Although it is trained for a more trustworthy RAG backbone model, we still recognize the potential risk of generating factually incorrect information (hallucinations).

\paragraph{License, Intended Use, and Sensitive Information.}
All data and software utilized in this study are derived from publicly available academic resources. Our primary datasets, QASPER and HotpotQA, are distributed under CC-BY-4.0 and CC-BY-SA-4.0 licenses, respectively, both of which permit use for academic research. These datasets contain no sensitive information, including unique identifiers of individual people or offensive content.
Furthermore, we leverage several open-source frameworks and baselines. The codebases for VeRL and LLaMA Factory are governed by the Apache 2.0 License, while ARENA is provided under the MIT License. For repositories where a specific license was not explicitly provided, we have used them strictly in accordance with their intended research purposes. All model checkpoints used are open-source weights hosted on Hugging Face; specifically, the TrustAlign series follows CC-BY-4.0, and the ARENA series follows the MIT License. Other models (SimpleDeepSearcher, R1-Searcher, and Search R1) were accessed via their public repositories.

\paragraph{Documentation of Artifacts.}
Regarding artifact documentation, our research primarily focuses on English. While the core tasks are English-centric, we observe incidental occurrences of other languages—such as Simplified Chinese and French—within certain Wikipedia passages in HotpotQA. These typically appear as localized names or references within the predominantly English text. In terms of domain coverage, QASPER spans diverse scientific literature, while HotpotQA provides broad coverage of general-world knowledge derived from Wikipedia. We have verified that these artifacts contain no sensitive demographic information and align with their original intended use for academic research.


\clearpage
\appendix
\section*{Appendix}

\section{Pseudo-code for the Reward Function}\label{app:pseudo-code}

\begin{algorithm}[H]
\caption{Pseudo-code for the reward function.}
\label{alg:reward}
\begin{algorithmic}[1]
\REQUIRE Model output $o$, ground-truth evidence set $e_{\mathrm{gt}}$, ground-truth answer set $a_{\mathrm{gt}}$
\ENSURE Reward $R(o)$

\STATE $y \gets$ extract the portion of $o$ after ``</think>''
\STATE $R_{\mathrm{f}} \gets \mathtt{CheckFormat}(y)$
\IF{$R_{\mathrm{f}} = 0$}
    \RETURN $0$
\ENDIF  \hfill \COMMENT{format violation → terminate}

\STATE $(R_{\mathrm{p}}, c) \gets \mathtt{CheckPath}(y)$
\IF{$R_{\mathrm{p}} = 0$}
    \RETURN $R_{\mathrm{f}}$ 
\ENDIF  \hfill \COMMENT{wrong path → terminate with format reward only}

\IF{$c = \texttt{<evidence>}$}
    \STATE $o_{\mathrm{evid}} \gets \mathtt{ExtractTag}(y, \texttt{<evidence>})$
    \STATE $o_{\mathrm{ans}}  \gets \mathtt{ExtractTag}(y, \texttt{<answer>})$
    \STATE $r_{\mathrm{e}} \gets \max\limits_{e \in e_{\mathrm{gt}}} \mathtt{ROUGE\_L}(o_{\mathrm{evid}}, e)$
    \STATE $r_{\mathrm{a}} \gets \max\limits_{a \in a_{\mathrm{gt}}} \mathtt{ROUGE\_L}(o_{\mathrm{ans}},  a)$
    \STATE $R_{\mathrm{c}} \gets \alpha r_{\mathrm{e}} + \beta r_{\mathrm{a}}$
    \RETURN $R_{\mathrm{f}} + R_{\mathrm{p}} + R_{\mathrm{c}}$
\ELSE
    \RETURN $R_{\mathrm{f}} + R_{\mathrm{p}}$ 
\ENDIF  \hfill \COMMENT{<llm> path → no content evaluation}

\end{algorithmic}
\end{algorithm}

\section{Detail Settings}\label{app:settings}

\subsection{Configurations}

To construct the training set, we employ diverse retrieval strategies with k=5: an oracle retriever, BM25 (k1=1.5,b=0.75), an embedding-based retriever (Qwen3-0.6B-Embedding), a retrieve-then-rerank pipeline (Qwen3-0.6B-Embedding + BGE-Reranker), and a random retriever.
For the training stage, VeRL (v0.4.1.dev0) serves as our training backend, with the hyperparameters shown in Table~\ref{tab:hyperparameters}. Each experimental run requires approximately 32 GPU hours (executed on 4 $\times$ NVIDIA A800-80G GPUs for 8 hours).

\begin{table}[tbp]
    \centering
    \caption{Main hyperparameters for our experiments.}
    \resizebox{!}{0.45\linewidth}{\begin{tabular}{c|c c}
    \toprule
    Parameter & Training & Inference \\
    \midrule
        clip\_low & 0.2 & N/A\\
        clip\_high & 0.28 & N/A\\
        clip\_ratio\_c & 10.0 & N/A\\
        learning\_rate & $2e-6$ & N/A\\
        training\_steps & 400 & N/A \\
        warmup\_steps & 50 & N/A\\
        weight\_decay & 0.1 & N/A\\
        grad\_clip & 1.0 & N/A\\
        max\_response\_length & 3072 & 3072 \\
        overlong\_buffer\_length & 1024 & N/A\\
        overlong\_penalty\_factor & 1 & N/A\\
        train\_prompt\_batchsize & 32 & N/A\\
        group\_size & 8 & N/A\\
        train\_prompt\_mini\_batchsize & 8 & N/A\\
        temperature & 1.0 & 0.6\\
        top\_p & 1.0 & 0.9\\
    \bottomrule
    \end{tabular}}
    \label{tab:hyperparameters}
\end{table}

For the baseline methods that required training, we utilize their corresponding training backends and training datasets in their code repositories, except for Trust-align, for which we employ LLaMA-Factory (0.9.4.dev0) to conduct DPO training based on the provided training datasets. 
All the baseline methods are adequately trained until the reward or loss converges. Other detailed hyperparameters are provided in the GitHub repositories.

\subsection{Prompt Templates}

For our training and inference, we adopt the same zero-shot prompt as Table~\ref{tab:our_prompt}.

\begin{table}[htb]
\centering
\caption{Prompt template for our method in both training and test stage.}
\phantomsection
\begin{tcolorbox}[colback=white!95!gray,colframe=gray!50!black,rounded corners,label={scale-depression}, title={Our Prompt}]
{\small System Prompt:

You are an evidence-validation assistant. For each query, you are given a question wrapped inside the <question>...</question> tag, and a series of documents as evidence wrapped inside the <ref>...</ref> tag. To solve these questions, you must follow this **exact** process without deviation:

\#\#\# Instructions:

1. You need to first figure out if the evidences are relevant and useful to the question. 

    - If the evidences are **NOT** relevant, you should recall your knowledge about this question and wrap the process of recalling inside the token of <llm>...</llm>. 

    - If the evidences are relevant, you should select the evidence that is most relevant to the question and wrap the selected texts inside the token of <evidence>...</evidence>. 

2. You should then answer the question based on the selected evidence or your knowledge. The answer should be wrapped inside the token of <answer>...</answer>.

User Prompt:

Here is the question and the references: 
<question>{question}</question>
<ref>
    {ref}
</ref>}

\end{tcolorbox}
\label{tab:our_prompt}
\end{table}

For the baselines, we adopt the corresponding prompt template with a manually constructed three-shot example from the corresponding training dataset. For example, as illustrated in Table~\ref{tab:3shot}, we sample three questions from the training set of HotpotQA and manually construct two of them as answerable and the other one as unanswerable. Similarly, to maintain distributional consistency, the three-shot examples for the QASPER test set were derived from its respective training split. Specifically, for baselines whose original prompt templates lacked an abstention instruction, we incorporated one with minimal intervention to ensure a fair comparison while preserving the core structure of the baseline.

\begin{table}[htb]
\centering
\caption{Three-shot examples for HotpotQA dataset.}
\phantomsection
\begin{tcolorbox}[colback=white!95!gray,colframe=gray!50!black,rounded corners,label={scale-depression}, title={3-shot Example}]
{\small User Prompt: **Question:**

Upper Denton is situated on the line of the Roman road that ran through the valleys of which two rivers?

**Context:**

...(With ground-truth evidence)...

Assistant: Tyne and Irthing.

User Prompt: **Question:**

In what year was the Enblish artist who released her second studio album Fall to Grace in 2012, born?

**Context:**

...(With ground-truth evidence)...

Assistant: 21 July 1981.

User Prompt: **Question:**

How many Tony Awards was the musical comedy on which Jonathan Tunick started working with Stephen Sondheim, nominated for ?

**Context:**

...(Without ground-truth evidence)...

Assistant: I don't know.}
\end{tcolorbox}
\label{tab:3shot}
\end{table}

\subsection{SFT Data Construction}\label{app:sft-data}

Following the data construction strategy described in Section~\ref{method:data}, we curate the input-output pairs for the SFT training set. While inputs remain consistent with the retrieved context format, the targets are synthetically generated using rule-based templates to enforce structural alignment with our reasoning paths. For answerable instances, the target output is formatted as: \texttt{<evidence>} \[GT Evidence\] \texttt{</evidence>} \texttt{<answer>} \[GT Answer\] \texttt{</answer>}. For unanswerable instances, the model is supervised to output: \texttt{<llm>} The question is unanswerable \texttt{</llm>}\texttt{<answer>} Unanswerable \texttt{</answer>}. We subsequently perform SFT using the LLaMA-Factory framework to initialize the model with basic instruction-following and formatting capabilities.

\section{Extended Experiments}\label{app:extended}

We provide extra analysis related to our experiments in this section.

\begin{table}[t]
\centering
\caption{Ablation study on HotpotQA, Qwen3-4B, with Qwen3-0.6B-Embedding (Qwen) or the BAAI-BGE-M3 (BAAI) as the retriever for the test set, where ER denotes \underline{E}vidence \underline{R}ouge-L score.}
\label{tab:ablation}
\resizebox{\linewidth}{!}{\begin{tabular}{lcccccgg}
\toprule
\multirow{3}{*}{} & \multicolumn{7}{c}{\textbf{HotpotQA - Dense Retriever Qwen}\small (296:204)} \\
\cmidrule(lr){2-8} 
& \multicolumn{4}{c}{Answerable} & {Unanswerable} & \multicolumn{2}{c}{\textbf{Overall}} \\
\cmidrule(lr){2-5} \cmidrule(lr){6-6} \cmidrule(lr){7-8}
& EM & F1 & LJ & ER & Acc. & \textbf{Acc.} & \textbf{B. Acc.}\\
\midrule
Naive & 66.89 & 86.51 & 91.22 & 45.43 & 59.80 & 78.40 & 75.51 \\
SR & 61.82 & 80.66 & 89.53 & 59.67 & 62.25 & 78.40 & 75.89 \\
W/O AB & 65.54 & 84.21 & 92.23 & 62.56 & 67.16 & 82.00 & 79.70\\
\midrule
W/O Weighting & 59.80 & 80.19 & 89.19 & 60.79 & 56.37 & 75.80 & 72.78 \\
EM Reward & 66.55 & 85.89 & 92.57 & 21.82 & 62.75 & 80.40 & 77.66 \\
\midrule
\ours & 65.20 & 84.95 & 91.55 & 70.36 & 66.18 & 81.20 & 78.87 \\
\midrule
\multirow{3}{*}{} & \multicolumn{7}{c}{\textbf{HotpotQA - Dense Retriever BAAI}\small (330:170)} \\
\midrule
W/O AB& 63.64 & 84.10 & 90.00 & 62.72 & 62.31 & 80.60 & 76.16 \\
\ours & 66.97 & 86.18 & 93.33 & 62.82 & 64.71 & 83.60 & 79.02 \\
\bottomrule
\end{tabular}}
\end{table}

\begin{table*}[ht]
\centering
\caption{Comparing results on QASPER Dataset among different retrieval settings, where $k=3$. The \textbf{best results} are bold, and the \underline{runner-up results} are underlined. The number of answerable and unanswerable questions in each dataset is denoted as \textbf{(\textit{answerable}: \textit{unanswerable})}. Methods with $\dagger$ sign come from the released checkpoints by the corresponding authors.}
\label{tab:qasper}
\resizebox{\textwidth}{!}{\begin{tabular}{lccccccccggcccccgg}
\toprule
\multirow{3}{*}{} & \multicolumn{3}{c}{\textbf{Oracle}\small(1451 : 0)} & \multicolumn{7}{c}{\textbf{BM25}\small(1074 : 377)} & \multicolumn{7}{c}{\textbf{Dense Retrieval \& Reranker}\small(1205 : 246)} \\
\cmidrule(lr){2-4} \cmidrule(lr){5-11} \cmidrule(lr){12-18} 
& \multicolumn{3}{c}{Answerable} & \multicolumn{3}{c}{Answerable} & \multicolumn{2}{c}{Unanswerable} & \multicolumn{2}{c}{\textbf{Overall}} & \multicolumn{3}{c}{Answerable} & \multicolumn{2}{c}{Unanswerable} & \multicolumn{2}{c}{\textbf{Overall}} \\
\cmidrule(lr){2-4} \cmidrule(lr){5-7} \cmidrule(lr){8-9} \cmidrule(lr){10-11} \cmidrule(lr){12-14} \cmidrule(lr){15-16} \cmidrule(lr){17-18}
& EM & F1 & LJ & EM & F1 & LJ & Acc. & LJ & \textbf{Acc.} & \textbf{B. Acc.} & EM & F1 & LJ & Acc. & LJ & \textbf{Acc.} & \textbf{B. Acc.} \\
\midrule
\rowcolor[HTML]{F2F2F2}
\multicolumn{18}{l}{\textbf{Prompt based}} \\
Qwen3-4B 
& 0.07 & 49.65 & \textbf{83.60} 
& 0.0  & 49.26 & \textbf{71.69} & 5.84 & 42.44 & 64.09 & 57.07
& 0.0  & 50.10 & \textbf{78.42} & 6.50 & 43.09 & \underline{72.43} & 60.76 \\
Llama3-8B-Instruct 
&  0.76 & 52.66 & 69.61 
&  1.12 & 52.92 & 64.62 & 16.71 & 34.22 & 56.72 & 49.42 
&  1.66 & 54.78 & 70.21 & 18.70 & 31.71 & 63.68 & 50.96\\
DeepSeek-chat 
& 0.14 & 62.64 & \underline{78.70} 
& 0.0  & 50.41 & 63.13 & 55.70 & 75.60 & \underline{66.37} & 69.37
& 0.08 & 51.96 & 69.63 & 44.31 & 69.92 & 69.68 & 69.78\\
SuRe on Qwen3-4B 
& 6.13 & 51.17 & 42.38
& 6.05 & 51.77 & 44.41 & 0.0 & 1.33 & 33.22 & 22.87
& 6.31 & 52.35 & 46.64 & 0.0 & 0.0  & 38.73 & 23.32\\
SuRe on Llama3-8B-Instruct
& 5.31 & 53.45 & 49.35 
& 4.75 & 54.23 & 55.68 & 0.0 & 0.27 & 41.90 & 29.17
& 5.23 & 54.50 & 55.27 & 0.0 & 1.63 & 46.18 & 28.45\\
\midrule
\rowcolor[HTML]{F2F2F2}
\multicolumn{18}{l}{\textbf{SFT based}} \\
Qwen3-4B 
& \underline{36.46} & \underline{76.61} & 74.64
& \textbf{26.91} & 66.26 & 55.21 & 71.09 & 71.35 & 59.41 & 63.28
& \textbf{30.71} & 69.75 & 62.57 & 70.73 & 70.73 & 63.96 & 66.65 \\
Llama3-8B-Instruct
& \textbf{38.04} & \textbf{77.34} & 74.50
& \underline{26.72} & 66.11 & 55.96 & 74.80 & 74.80 & 60.85 & 65.38
& \underline{27.88} & 70.09 & 63.24 & 67.89 & 67.89 & 64.02 & 65.57\\
SimpleDeepSearcher-7B 
& 23.23 & 65.91 & 71.47
& 22.07 & 65.15 & 63.13 & 18.83 & 19.89 & 51.90 & 41.51
& 22.74 & 66.57 & 65.64 & 20.33 & 21.54 & 58.17 & 43.59 \\
\midrule
\rowcolor[HTML]{F2F2F2}
\multicolumn{18}{l}{\textbf{RL based}} \\
TrustAlign-3B $^\dagger$
& 0.0 & 42.48 & 50.24 
& 0.0 & 40.54 & 39.85 & 74.27 & 78.51 & 49.90 & 59.18
& 0.0 & 41.99 & 42.74 & 71.54 & 76.42 & 48.45 & 59.58\\
TrustAlign on Qwen3-4B
& 0.0 & 44.05 & 47.48 
& 0.0 & 43.53 & 31.28 & 0.0   & 66.05 & 40.32 & 48.67
& 0.0 & 44.03 & 37.76 & 0.0   & 63.41 & 42.11 & 50.59\\
TrustAlign-7B $^\dagger$
& 0.0 & 44.19 & 47.07 
& 0.0 & 43.03 & 32.68 & 0.0   & \underline{87.80} & 47.00 & 60.24
& 0.0 & 43.92 & 35.93 & 0.0   & \underline{85.37} & 44.31 & 40.12 \\
TrustAlign-8B $^\dagger$
& 0.0 & 35.07 & 17.99 
& 0.0 & 34.42 & 8.66  & \textbf{95.76} & \textbf{96.02} & 31.36 & 52.34
& 0.0 & 35.16 & 12.53 & \textbf{97.97} & \textbf{97.97} & 27.02 & 55.25\\
ARENA on Qwen3-4B 
& 24.19 & 66.43 & 76.15 
& 23.74 & 67.24 & \underline{68.81} & 23.08 & 23.08 & 56.93 & 49.95
& 25.14 & 68.77 & \underline{73.86} & 21.54 & 21.95 & 65.06 & 47.91\\
ARENA-7B $^\dagger$
& 23.64 & 68.73 & 76.91 
& 21.60 & 68.01 & 67.32 & 6.90 & 8.75 & 52.10 & 38.04
& 23.49 & 69.99 & 69.71 & 6.91 & 8.13 & 59.27 & 38.92\\
ARENA-8B $^\dagger$
& 24.33 & 68.76 & 68.92 
& 22.81 & 67.09 & 66.29 & 0.0 & 0.0 & 49.07 & 33.15
& 22.82 & 69.01 & 70.12 & 0.0 & 0.0 & 58.24 & 35.06\\
R1 Searcher-7B$^\dagger$
& 15.02 & 57.49 & 48.04
& 14.34 & 58.27 & 43.20 & 13.26 & 16.98 & 36.39 & 30.09 
& 16.85 & 60.49 & 47.88 & 14.23 & 17.48 & 42.73 & 32.68 \\
Search R1-7B$^\dagger$
& 20.19 & 63.91 & 64.78
& 18.16 & 62.58 & 56.61 & 32.36 & 34.48 & 50.86 & 45.55
& 19.75 & 65.34 & 61.74 & 32.52 & 35.77 & 57.34 & 48.76 \\
\midrule
\rowcolor[HTML]{F2F2F2}
\multicolumn{18}{l}{\textbf{GRACE}} \\
Qwen3-4B 
& 29.84 & 73.11 & 76.36 
& 24.39 & \textbf{70.01} & 68.25 & 77.19 & 77.19 & \textbf{70.57} & \textbf{72.72}
& 27.22 & \textbf{72.73} & 73.78 & 77.24 & 77.24 & \textbf{74.36} & \textbf{75.51}\\
Llama3-8B-Instruct
& 29.98 & 73.42 & 76.91 
& 23.93 & \underline{68.44} & 59.12 & \underline{83.02} & 83.02 & 65.33 & \underline{71.07}
& 27.47 & \underline{71.13} & 67.14 & \underline{78.86} & 78.86 & 69.12 & \underline{73.14} \\
\bottomrule
\end{tabular}}
\end{table*}

\begin{table}[th]
\centering
\caption{Comparing results on HotpotQA Dataset among different retrieval settings.}
\label{tab:topk_transposed}
\resizebox{0.95\linewidth}{!}{
\begin{tabular}{cl ccc cc cc}
\toprule
\multirow{2}{*}{\textbf{Method}} & \multirow{2}{*}{\textbf{$k$}} & \multicolumn{3}{c}{\textbf{Answerable}} & \multicolumn{2}{c}{\textbf{Unanswerable}} & \multicolumn{2}{c}{\textbf{Overall}} \\
\cmidrule(lr){3-5} \cmidrule(lr){6-7} \cmidrule(lr){8-9}
 & & EM & F1 & LJ & Acc. & LJ & Acc. & B. Acc. \\
\midrule

\multirow{4}{*}{BM25} 
 & 3 & 68.20 & 86.53 & 93.55 & 61.13 & 61.13 & 75.20 & 77.34 \\
 & 4 & 68.25 & 87.47 & 91.61 & 68.14 & 68.14 & 81.00 & 79.88 \\
 & 5 & 67.51 & 86.88 & 91.17 & 69.94 & 69.94 & 83.40 & 80.56 \\
 & 6 & 66.67 & 87.03 & 90.20 & 67.83 & 67.83 & 83.80 & 79.02 \\
\midrule

\multirow{4}{*}{\shortstack[c]{Dense\\Retriever}} 
 & 3 & 65.20 & 84.95 & 91.55 & 66.18 & 66.18 & 81.20 & 78.87 \\
 & 4 & 68.56 & 87.21 & 90.42 & 74.10 & 74.10 & 85.00 & 82.26 \\
 & 5 & 68.31 & 86.92 & 90.44 & 70.90 & 70.90 & 85.20 & 80.67 \\
 & 6 & 66.50 & 86.43 & 89.24 & 72.53 & 72.53 & 86.20 & 80.89 \\
\midrule

\multirow{4}{*}{\shortstack[c]{Retriever\&\\Reranker}} 
 & 3 & 66.59 & 86.10 & 92.01 & 62.07 & 62.07 & 86.90 & 77.04 \\
 & 4 & 69.40 & 87.97 & 92.46 & 71.43 & 71.43 & 90.40 & 81.95 \\
 & 5 & 67.72 & 87.64 & 90.72 & 50.00 & 50.00 & 88.60 & 70.36 \\
 & 6 & 66.88 & 87.06 & 90.42 & 50.00 & 50.00 & 88.80 & 70.21 \\

\bottomrule
\end{tabular}
}
\end{table}

\begin{table*}[ht]
\centering
\caption{Comparing results on HotpotQA Dataset among different retrieval settings, where $k=3$. The \textbf{best results} are bold, and the \underline{runner-up results} are underlined. The number of answerable and unanswerable questions in each dataset is denoted as \textbf{(\textit{answerable}: \textit{unanswerable})}. Methods with $\dagger$ sign come from the released checkpoints by the corresponding authors.}
\label{tab:hotpot}
\resizebox{\textwidth}{!}{\begin{tabular}{lccccccccggcccccgg}
\toprule
\multirow{3}{*}{} & \multicolumn{3}{c}{\textbf{Oracle}\small(500 : 0)} & \multicolumn{7}{c}{\textbf{BM25}\small(217 : 283)} & \multicolumn{7}{c}{\textbf{Dense Retrieval \& Reranker}\small(413 : 87)} \\
\cmidrule(lr){2-4} \cmidrule(lr){5-11} \cmidrule(lr){12-18} 
& \multicolumn{3}{c}{Answerable} & \multicolumn{3}{c}{Answerable} & \multicolumn{2}{c}{Unanswerable} & \multicolumn{2}{c}{\textbf{Overall}} & \multicolumn{3}{c}{Answerable} & \multicolumn{2}{c}{Unanswerable} & \multicolumn{2}{c}{\textbf{Overall}} \\
\cmidrule(lr){2-4} \cmidrule(lr){5-7} \cmidrule(lr){8-9} \cmidrule(lr){10-11} \cmidrule(lr){12-14} \cmidrule(lr){15-16} \cmidrule(lr){17-18}
& EM & F1 & LJ & EM & F1 & LJ & Acc. & LJ & \textbf{Acc.} & \textbf{B. Acc.} & EM & F1 & LJ & Acc. & LJ & \textbf{Acc.} & \textbf{B. Acc.} \\
\midrule
\rowcolor[HTML]{F2F2F2}
\multicolumn{18}{l}{\textbf{Prompt based}} \\
Qwen3-4B 
& 23.80 & 58.73 & \underline{91.60}
& 13.36 & 52.52 & 92.17 & 8.13 & 57.24 & 72.40 & \underline{74.71}
& 19.13 & 55.47 & \underline{90.31} & 8.05 & 59.77 & \underline{85.00} & \underline{75.04}\\
Llama3-8B-Instruct 
& 22.80 & 50.47 & 43.20
&  6.45 & 37.73 & 28.11 & 42.76 & 87.28 & 61.60 & 57.70
&  6.78 & 37.69 & 24.21 & 42.53 & 86.21 & 35.00 & 55.21\\
DeepSeek-chat 
& 44.20 & 67.93 & 76.20 
& 37.79 & 63.65 & 72.35 & \underline{76.68} & 76.68 & \underline{74.80} & 74.52
& 36.80 & 63.05 & 72.15 & \underline{70.11} & 70.11 & 71.80 & 71.13\\
SuRe on Qwen3-4B 
& 29.80 & 65.79 & 67.00
& 25.35 & 63.54 & 67.74 & 0.0 & 0.0 & 29.40 & 33.87
& 29.30 & 65.65 & 69.25 & 0.0 & 0.0 & 57.20 & 34.62\\
SuRe on Llama3-8B-Instruct 
& 37.00 & 69.93 & 82.20
& 33.64 & 67.62 & 79.26 & 0.0 & 1.06 & 35.00 & 40.16
& 35.59 & 68.77 & 77.97 & 0.0 & 0.0 & 64.40 & 38.99\\
\midrule
\rowcolor[HTML]{F2F2F2}
\multicolumn{18}{l}{\textbf{SFT based}} \\
Qwen3-4B 
& 66.00 & 86.20 & 87.20
& 64.98 & \underline{85.87} & 86.18 & 50.18 & 50.18 & 65.80 & 68.18
& \underline{67.07} & 85.95 & 85.71 & 36.78 & 36.78 & 77.20 & 61.25\\
Llama3-8B-Instruct 
& \textbf{67.40} & \textbf{87.13} & 88.00
& \underline{65.44} & 85.69 & 85.71 & 54.06 & 54.06 & 68.20 & 69.89
& \textbf{68.52} & \underline{86.82} & 87.17 & 40.23 & 40.23 & 79.00 & 63.70\\
SimpleDeepSearcher-7B$^\dagger$
& 40.40 & 65.13 & 60.60
& 43.78 & 68.34 & 65.44 & 65.02 & 65.02 & 65.20 & 65.23
& 43.34 & 67.81 & 60.77 & 55.17 & 55.17 & 59.80 & 57.97 \\
\midrule
\rowcolor[HTML]{F2F2F2}
\multicolumn{18}{l}{\textbf{RL based}} \\
TrustAlign-3B$^\dagger$ 
& 0.00 & 43.35 & 61.00 
& 0.00 & 40.88 & 50.69 & 53.71 & 71.38 & 62.40 & 61.04
& 0.00 & 39.92 & 49.64 & 59.77 & 68.97 & 53.00 & 59.31\\
TrustAlign on Qwen3-4B
& 0.0 & 32.91 & 21.20 
& 0.0 & 32.55 & 19.82 & 0.0   & \textbf{98.23} & 64.20 & 59.03
& 0.0 & 31.97 & 12.11 & 0.0   & \textbf{97.70} & 27.00 & 54.91\\
TrustAlign-7B$^\dagger$ 
& 0.00 & 36.55 & 36.60 
& 0.00 & 34.55 & 30.88 & 0.00 & 90.46 & 64.60 & 60.67
& 0.00 & 34.48 & 27.12 & 0.00 & 87.36 & 37.60 & 57.24\\
TrustAlign-8B$^\dagger$ 
& 0.00 & 34.75 & 34.00 
& 0.00 & 32.35 & 28.57 & \textbf{90.81} & \underline{90.81} & 63.80 & 59.69
& 0.00 & 31.88 & 25.18 & \textbf{93.10} & \underline{93.10} & 37.00 & 59.14\\
ARENA on Qwen3-4B
& 63.40 & 85.57 & 89.80
& 63.59 & 85.54 & \underline{92.63} & 42.05 & 42.40 & 64.20 & 67.52
& 64.16 & 85.62 & 89.83 & 37.93 & 37.93 & 80.80 & 63.88\\
ARENA-7B$^\dagger$ 
& 66.20 & 86.04 & 90.20 
& 64.06 & 85.81 & 91.24 & 6.01  & 13.07 & 47.00 & 52.16
& 64.89 & 86.01 & 88.14 & 5.75  & 11.49 & 74.80 & 49.82\\
ARENA-8B$^\dagger$
& 62.00 & 84.34 & 86.80 
& 58.06 & 81.70 & 85.25 & 0.00 & 0.00 & 37.00 & 42.63
& 62.95 & 84.41 & 86.92 & 0.00 & 0.00 & {71.80} & 43.46\\
R1 Searcher-7B$^\dagger$
& 45.00 & 68.32 & 62.40
& 47.00 & 73.40 & 69.59 & 55.83 & 55.83 & 61.80 & 62.71
& 49.88 & 73.57 & 68.77 & 63.22 & 63.22 & 67.80 & 65.99 \\
Search R1-7B$^\dagger$
& 55.20 & 79.12 & 79.20
& 56.22 & 79.42 & 81.11 & 44.17 & 44.17 & 60.20 & 62.64
& 55.45 & 78.86 & 75.30 & 37.93 & 40.23 & 69.20 & 57.77 \\
\midrule
\rowcolor[HTML]{F2F2F2}
\textbf{GRACE} & & & & & & & & & & & & & && & &\\
Qwen3-4B 
& \underline{66.60} & \underline{86.30} & \textbf{92.00} 
& \textbf{68.20} & \textbf{86.53} & \textbf{93.55} & 61.13 & 61.13 & \textbf{75.20} & \textbf{77.34}
& 66.59 & 86.10 & \textbf{92.01} & 62.07 & 62.07 & \textbf{86.90} & \textbf{77.04}\\
Llama3-8B-Instruct
& 65.60 & 86.21 & 88.20 
& 62.67 & 85.65 & 86.18 & 47.35 & 50.18 & 65.80 & 68.18
& 66.83 & \textbf{87.03} & 88.38 & 31.03 & 34.48 & 79.00 & 61.43 \\
\bottomrule
\end{tabular}}
\end{table*}

\subsection{Ablation Study}\label{app:ablation-study}

To verify our proposed pipeline, we conduct an ablation study along two axes: data construction and reward design.
On the data side, we consider three variants.
(1) \textbf{Naive Method}. As stated in Section~\ref{method:data}, we use this naive approach to construct the training data.
(2) \textbf{Single-retriever} (SR). To verify the effectiveness of heterogeneous retrievers, we only keep training data constructed by the embedding-based retriever Qwen3-0.6B-Embedding and train for the same number of steps, denoted as single-retriever.
(3) \textbf{Without class balancing} (W/O CB). 
To verify the importance of class balancing, we disable our retrieved passages manipulation component.
On the reward side, we ablate two design choices.
(1) \textbf{Without evidence-answer weighting} (W/O weighting). We remove the asymmetric weighting between evidence grounding and answer accuracy, i.e., $\alpha$ and $\beta$, letting them contribute equally to the objective.
(2) \textbf{Exact-match reward} (EM Reward). We replace the Rouge-L-based reward with an exact-match-based reward for both evidence and answer supervision.

    

As illustrated in Table~\ref{tab:ablation}, \ours\ strikes an optimal balance between performance and generalization.
From the data perspective, adopting the naive approach for training data construction leads to a significant performance degradation: overall accuracy drops by approximately 3\%, while evidence grounding plummets by nearly 25\%. These results underscore the inefficacy of the naive construction method in capturing complex reasoning requirements.
For the single retriever variant, we observed that the model still acquires some knowledge. However, its overall performance is significantly lower than that achieved with heterogeneous retrievers, which may be attributed to the lack of data variance. 
Regarding retrieved passages manipulation, we found that removing this component leads to a slight performance gain—but only if the retriever used during testing is the same one used for data construction. 
Furthermore, evaluations on unseen embedding models indicate that retrieved passage manipulation primarily enhances generalization, allowing the model to maintain robust performance across different retrievers.

From the reward-design perspective, evidence-answer weighting emphasizes the importance of producing correct answers and well-grounded evidence, thereby improving both answer quality and evidence grounding. Moreover, our Rouge-L-based evidence reward provides a dense supervision signal: for evidence sentences that are difficult to match exactly, it yields graded and discriminative rewards rather than a binary signal, which ensures that the model receives meaningful feedback even for partially correct extractions, thereby facilitating stable convergence and consistent optimization.

\subsection{Performance on Different Retriever}\label{app:retriever}


In our main experiments, we fix the Qwen3-0.6B-Embedding as our retriever. To test the generalization of our method, we also evaluate it against all baselines on three different retriever settings: oracle, which directly inputs the ground truth evidence chunks; BM25; and the Retriever Reranker pipeline. The results on the QASPER and HotpotQA are illustrated in Table~\ref{tab:qasper} and Table~\ref{tab:hotpot}, respectively. For the oracle retriever, even if the ground truth evidence is input, the model still cannot answer all the questions correctly, which indicates the necessity of research on improving the model's capability of leveraging evidence. For the BM25 as the retriever, the ratio of correctly answering the answerable questions decreases compared to the oracle retriever and dense retriever. We attributed this phenomenon to the lower retrieval quality of BM25, which may provide unrelated evidence to the model with a higher score.  For the dense retriever with a reranker, the results are better than the BM25, demonstrating that the quality of the retriever matters for the RAG system, but with proper training, models with a relatively weak retriever can outperform models with a stronger one. For example, our method with dense retriever outperforms ARENA-4B with the oracle retriever, demonstrating the importance of this research area.

\subsection{Top-k Variants}\label{app:top-k}

In our main experiments, we fix $k{=}3$, but practical RAG systems often operate with dynamic or task-specific $k$.
As illustrated in Table~\ref{tab:topk_transposed}, our method is robust to the choice of $k$, with balanced accuracy remaining stable at around 80 across most settings. The only notable exception is the Dense Retriever \& Reranker configuration: as $k$ increases, the retriever almost always returns key evidence for each query, leaving very few unanswerable cases. This substantially reduces the effective sample size for abstention evaluation, leading to higher variance and larger fluctuations in performance.

\begin{table}[t]
\centering
\small
\setlength{\tabcolsep}{6pt}
\caption{Cross-dataset generalization results.}
\resizebox{\linewidth}{!}{\begin{tabular}{lccccccc}
\toprule
& \multicolumn{3}{c}{Answerable} & \multicolumn{2}{c}{Unanswerable} & \multicolumn{2}{c}{Overall} \\
Retriever & EM & F1 & LJ & Acc. & LJ & Acc. & B. Acc. \\
\midrule
Oracle  & 62.40 & 85.50 & 90.00 & -- & -- & -- & -- \\
Dense Retriever & 64.53 & 85.15 & 90.88 & 59.31 & 59.31 & 78.00 & 75.10 \\
BM25            & 59.91 & 84.35 & 90.32 & 60.07 & 60.07 & 73.20 & 75.20\\
Retriever \& Reranker  & 63.92 & 85.20 & 89.59 & 55.17 & 55.17 & 83.60 & 72.38 \\
\bottomrule
\end{tabular}}
\label{tab:cross_dataset_hotpot}
\end{table}

\subsection{Performance on Out-of-distribution Data}\label{app:ood}

As shown in Table~\ref{tab:cross_dataset_hotpot}, our model remains robust under cross-dataset evaluation: when trained on a different dataset, it generalizes well to the target dataset. Relative to the in-domain trained model, it shows an average drop of 2 percentage points in overall accuracy across target settings, while still outperforming the 3-shot untrained model.


\subsection{Impact on General Capabilities}\label{app:general}

We further evaluate the post-training model on a subset of MMLU, including abstract algebra, anatomy, computer security, econometrics, global facts, high school biology, high school chemistry, high school physics, human sexuality, professional law, and U.S. foreign policy, to assess whether RLVR introduces degradation in general knowledge and reasoning ability. Our results demonstrate that the proposed method preserves, and in some cases enhances, the model's general reasoning capabilities. Specifically, while the untrained Qwen3-4B base model achieves a score of 66.0\% on MMLU, our trained versions—Grace-Qwen3-4B-HotpotQA and Grace-Qwen3-4B-QASPER—attain 66.1\% and 67.6\% respectively, indicating that the specialized training does not compromise foundational knowledge. This result aligns with prior findings: RLVR updates only a small but functionally critical subset of parameters, preserving the pretrained model’s general competence~\cite{rlvr-small-set}, and the on-policy training mitigates catastrophic forgetting~\cite{rlvr-onpolicy}.

\begin{figure}[t]
    \centering
    \includegraphics[width=0.85\linewidth]{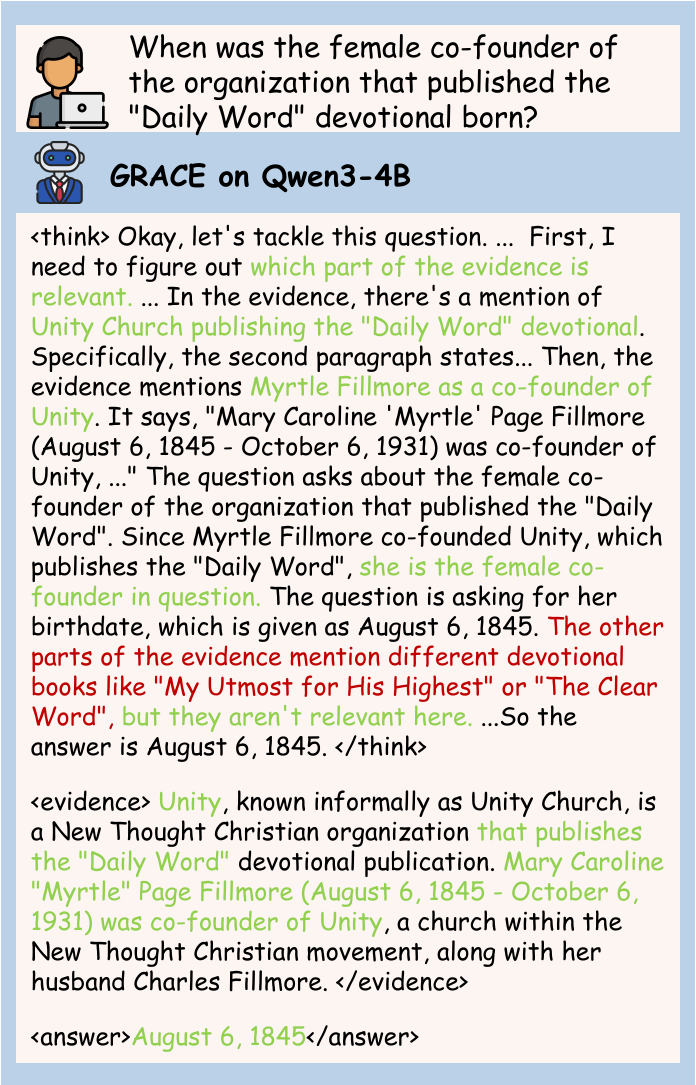}
    \caption{Case study on Grace: evidence-path selection with Qwen3-4B. Green highlights the model’s key reasoning steps, while red marks incorrect attempts.}
    \label{fig:qwen-evi}
\end{figure}

\begin{figure}[t]
    \centering
    \includegraphics[width=0.85\linewidth]{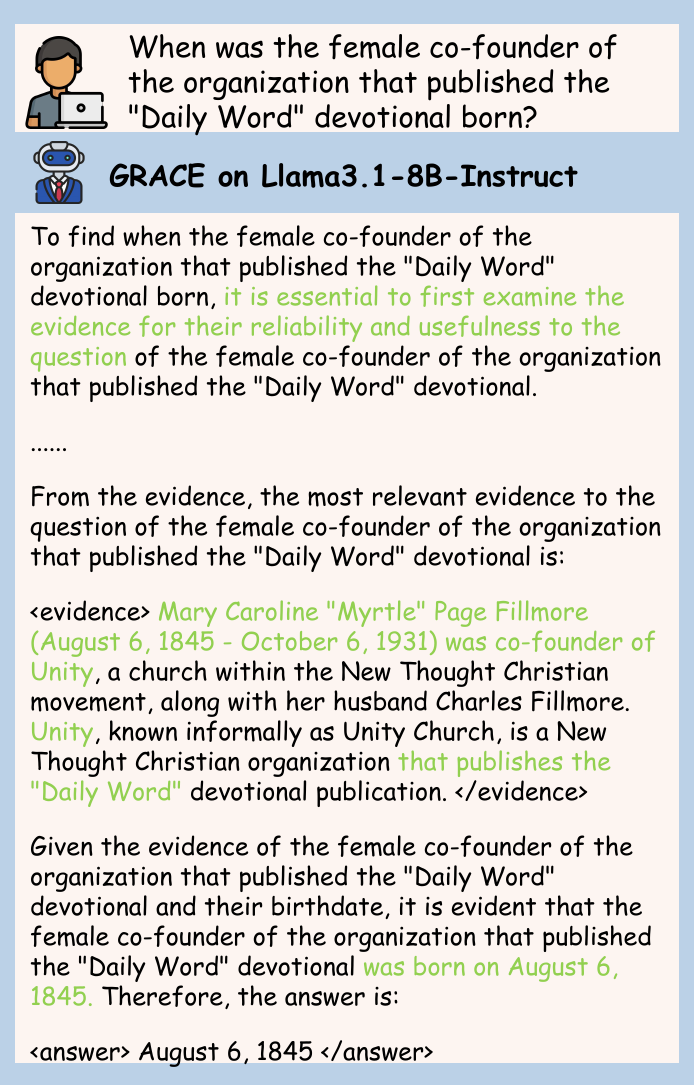}
    \caption{Case study on Grace: evidence-path selection with Llama3.1-8B-Instruct. Green highlights the model’s key reasoning steps.}
    \label{fig:llama-evi}
\end{figure}

\begin{figure}[t]
    \centering
    \includegraphics[width=0.85\linewidth]{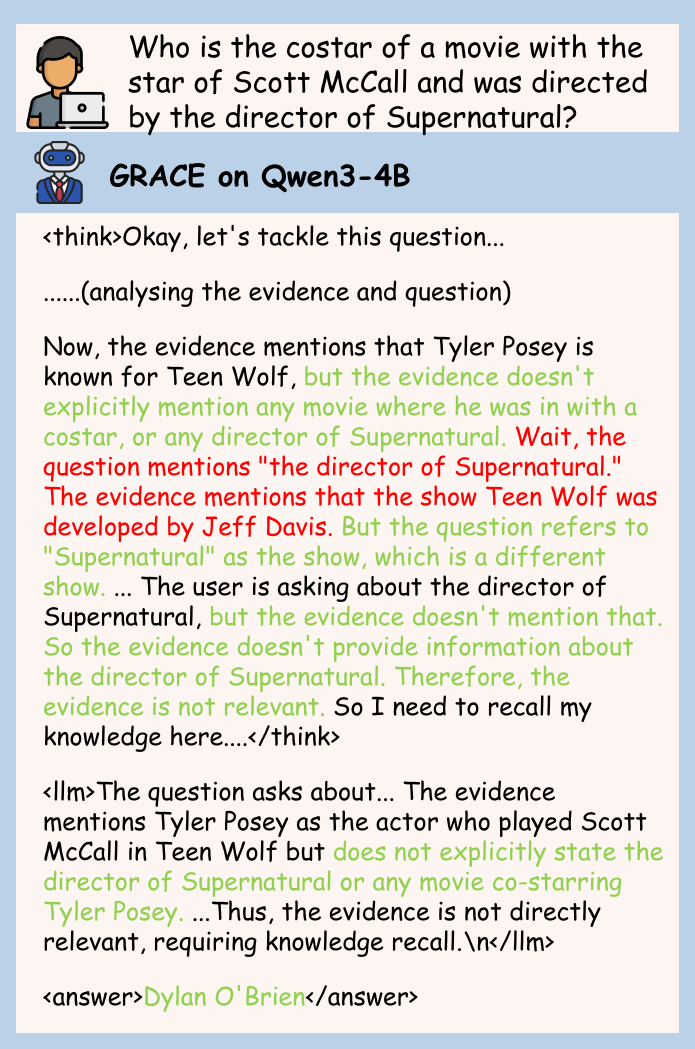}
    \caption{Case study on Grace: llm-path selection with Qwen3-4B. Green highlights the model’s key reasoning steps, while red marks incorrect attempts.}
    \label{fig:qwen-llm}
\end{figure}

\begin{figure}[t]
    \centering
    \includegraphics[width=0.85\linewidth]{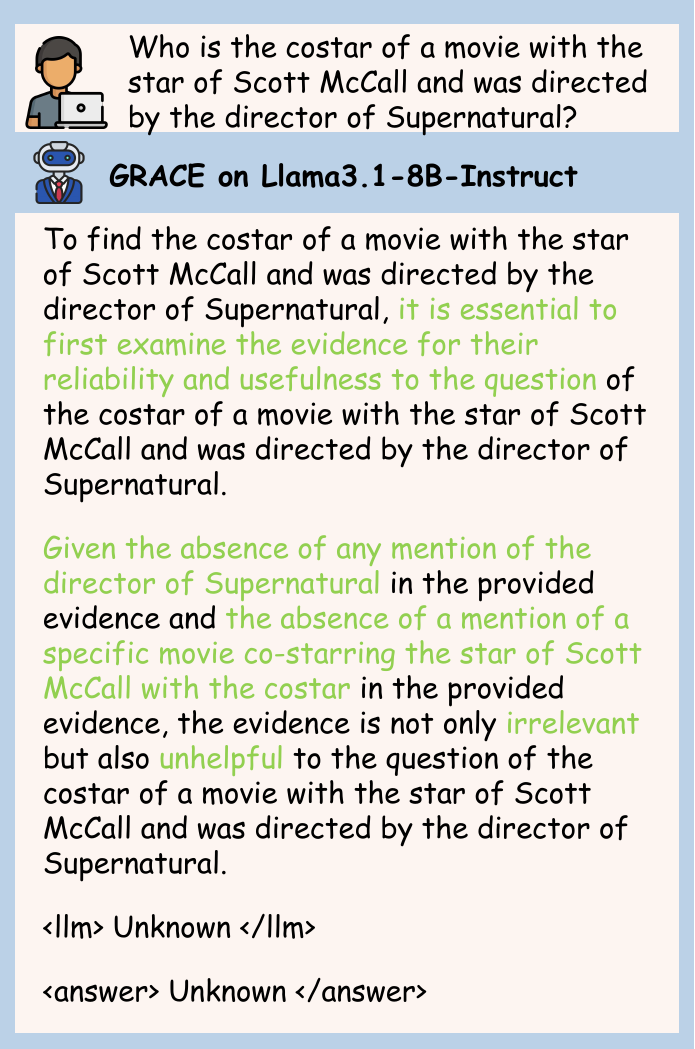}
    \caption{Case study on Grace: llm-path selection with Llama3.1-8B-Instruct. Green highlights the model’s key reasoning steps.}
    \label{fig:llama-llm}
\end{figure}

\subsection{Case Study}\label{app:case-study}

To better illustrate our models’ behavior, we present several examples from the trained models. As shown in Figures~\ref{fig:qwen-evi} and~\ref{fig:llama-evi}, the trained models learn to verify the provided evidence and perform intermediate reasoning before producing the final evidence selection and answer. 
Notably, even the instruct model, under our reward design, spontaneously acquires an evidence-checking procedure that resembles an explicit ``think''-style process.
Regarding the LLM path, we observe that the model attempts to retrieve relevant information from its internal parametric knowledge. As illustrated in Figure~\ref{fig:qwen-llm}, the trained model identifies that the provided evidence is insufficient, successfully recalls the necessary knowledge, and answers the question correctly. In contrast, for the relatively weaker Llama model (Figure~\ref{fig:llama-llm}), the trained model lacks the relevant knowledge and consequently responds with ``unknown''.

\section{The Use of LLMs}

This paper employed LLMs solely for grammatical correction and stylistic refinement, with the purpose of more effectively communicating our results and conclusions.

\end{document}